\documentclass[10pt,twocolumn,letterpaper]{article}

\usepackage[pagenumbers]{cvpr} 

\usepackage{multirow}
\usepackage{algorithmicx,algorithm}
\usepackage{algpseudocode}

\definecolor{cvprblue}{rgb}{0.21,0.49,0.74}
\usepackage[pagebackref,breaklinks,colorlinks,allcolors=cvprblue]{hyperref}

\def\paperID{} 
\def\confName{}
\def\confYear{}

\title{FreeArtGS: Articulated Gaussian Splatting Under Free-moving Scenario}

\author{
Hang Dai$^{1,3*}$ \quad
Hongwei Fan$^{1,3*}$ \quad
Han Zhang$^{2,3*}$ \quad
Duojin Wu$^{1,3}$ \quad
Jiyao Zhang$^{1,3}$ \quad
Hao Dong$^{1,3\dag}$ \\
$^{1}$CFCS, School of Computer Science, Peking University \quad \\
$^{2}$College of Computer Science and Technology, Zhejiang University \quad
$^{3}$PrimeBot}

\begin{document}
\maketitle
\begin{abstract}
The increasing demand for augmented reality and robotics is driving the need for articulated object reconstruction with high scalability. 
However, existing settings for reconstructing from discrete articulation states or casual monocular videos require non-trivial axis alignment or suffer from insufficient coverage, limiting their applicability. 
In this paper, we introduce FreeArtGS, a novel method for reconstructing articulated objects under free-moving scenario, a new setting with a simple setup and high scalability. 
FreeArtGS combines free-moving part segmentation with joint estimation and end-to-end optimization, taking only a monocular RGB-D video as input. 
By optimizing with the priors from off-the-shelf point-tracking and feature models, the free-moving part segmentation module identifies rigid parts from relative motion under unconstrained capture. The joint estimation module calibrates the unified object-to-camera poses and recovers joint type and axis robustly from part segmentation. Finally, 3DGS-based end-to-end optimization is implemented to jointly reconstruct visual textures, geometry, and joint angles of the articulated object. We conduct experiments on two benchmarks and real-world free-moving articulated objects. Experimental results demonstrate that FreeArtGS consistently excels in reconstructing free-moving articulated objects and remains highly competitive in previous reconstruction settings, proving itself a practical and effective solution for realistic asset generation. The project page is available at: \url{https://freeartgs.github.io/}
\end{abstract}  
\renewcommand{\thefootnote}{}
\footnote{*: Equal contributions. $\dagger$: Corresponding author.}

\section{Introduction}

Articulated objects broadly exist and are frequently interacted with in our daily lives. Building digital replicas of interactable articulated objects not only enhances the human experience in augmented reality~\cite{xia2025drawer}, but also reduces the sim-to-real gap~\cite{kerrrobot,yu2025artgs,jin2025artvip,weng2024neural,chen2024urdformer} for robot learning. To efficiently expand the scope of graphics- and simulation-ready assets, the reconstruction system for articulated objects should achieve \textit{high scalability} in a \textit{simple setup}.

\begin{figure}[t]
    \centering
    \includegraphics[width=1.0\linewidth]{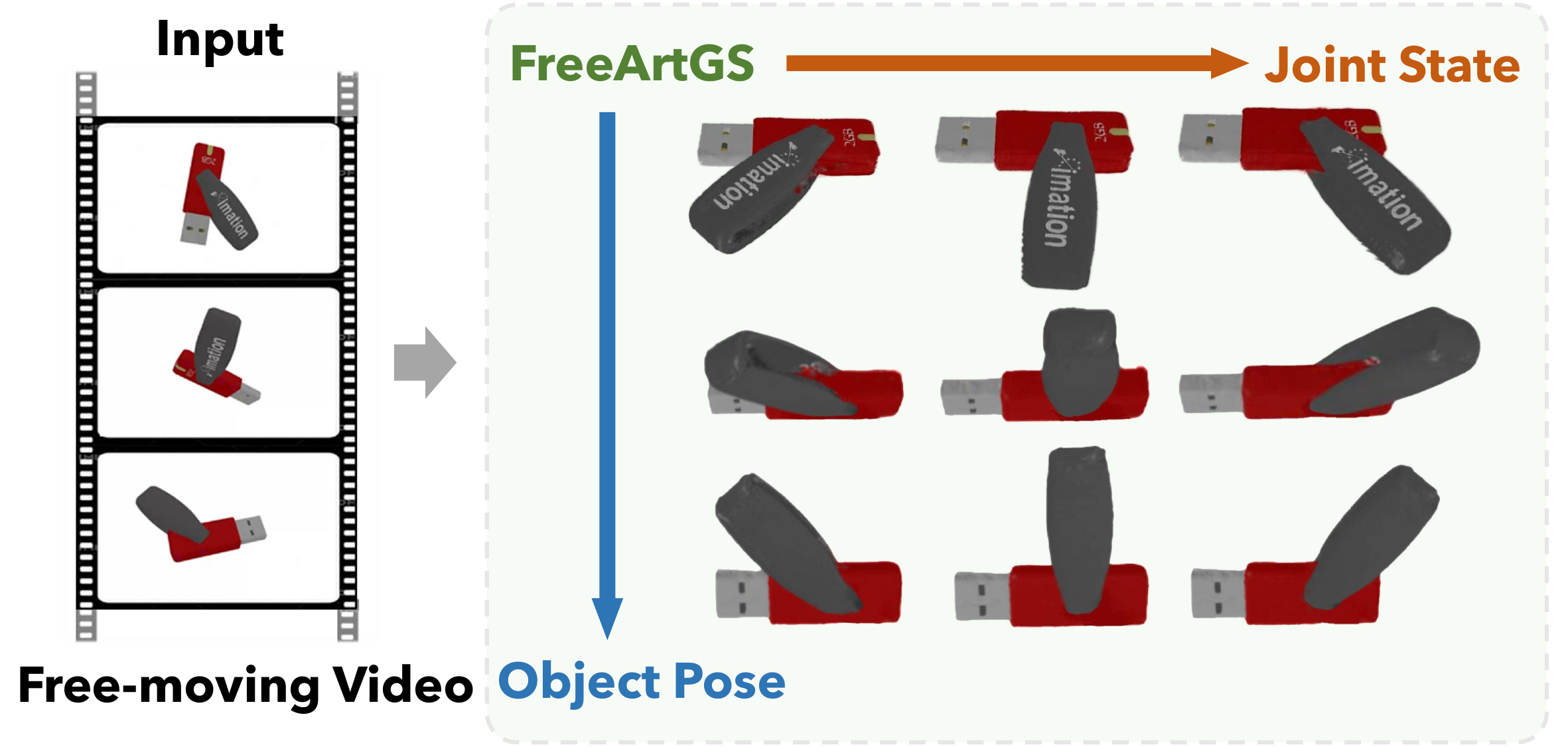}
    \caption{FreeArtGS reconstructs articulated object under the free-moving scenario, in which both joint and object pose move in an unconstrained manner.}
    \label{fig:1}
\vspace{-15pt}
\end{figure}

Regarding the scalability and simplicity of reconstructing articulated objects, recent works can be separated into three lines. 
The first line of works~\cite{articulateanything,realcode,ditto} directly generates the articulated object assets from a single-view image with foundation models. These methods fail to generalize to unseen scenarios due to the scarcity of post-optimization.
The second line of works~\cite{artgs,reartgs,splart} assumes that the object is captured in two articulation states (e.g., open-closed, pulled-pushed) with fixed multi-view cameras. However, these methods require alignment of the axes between the two states, limiting the real-world usage. 
The third line of works~\cite{rsrd,monomobility,video2articulation} reconstructs the object from casual monocular video with a static base part and moving dynamic parts. These works expose two disadvantages. First, during real-world casual capture, the pose of many articulated objects, such as scissors and pliers, may be inadvertently altered, violating the assumption of a static base part. Second, the coverage of the object is insufficient, limiting its usage in asset generation. These drawbacks motivate the need for reconstructing articulated objects in \textit{free-moving scenario}, in which both \textbf{joint state} and \textbf{object pose} relative to the camera vary arbitrarily (Fig.~\ref{fig:1}). By reconstructing the articulated object in a free-moving manner, the object is captured with full coverage of both object pose and joint state, resulting in an interactable, simulation-ready asset.


In this paper, we introduce \textbf{FreeArtGS}, 
a reconstruction system for articulated objects under the free-moving scenario. 
Our system consists of three key modules: 
\textbf{(1) Free-moving Part Segmentation.} We design an optimization-based part segmentation method based on the visual cues from dense 2D tracks~\cite{alltracker} and a pretrained feature extractor~\cite{simeoni2025dinov3}, without assuming a static base part or predefined motion patterns.
\textbf{(2) Joint Estimation.} We estimate the articulated joints by predicting the part-to-camera transformations, and use the relative transformation between the parts to decide the joint type and axis.
\textbf{(3) End-to-end Optimization.} We jointly refine the appearance, geometry, cameras, and articulation in an end-to-end manner, with ground-truth RGB, depth, and foreground masks as supervision.


To evaluate the reconstruction quality of our method under the free-moving scenario, we establish a new benchmark, FreeArt-21, including 21 free-moving articulated objects from 7 categories in the PartNet-Mobility dataset~\cite{partnetmobility}. We capture the free-moving object in the simulation engine Sapien~\cite{partnetmobility}, and tele-operate the object poses and joint states with a VR system. We also evaluate the method on real-world objects captured by an RGB-D camera. To align the settings of the existing baselines, we further compare our method with them in the Video2Articulation-S dataset. In all three evaluation settings, our method outperforms the current baselines by a large margin. 



To summarize, our contributions are threefold:

\begin{itemize}
    \item We propose FreeArtGS, a system for reconstructing articulated objects in free-moving scenarios, where the joint state and object pose vary arbitrarily without any static base part as a reference. Our approach combines motion-based part segmentation with joint estimation and end-to-end Gaussian Splatting optimization, enabling accurate reconstruction from only a monocular RGB-D video.
    \item Since there is no previous benchmark on articulated object reconstruction in the free-moving scenario, to bridge this gap, we build FreeArt-21, a simulated benchmark covering 21 free-moving articulated objects from 7 categories. To mimic the free-moving setting as in the real world, we develop a VR system to manipulate the object pose and joint state in the Sapien~\cite{partnetmobility} simulator.
    \item Experiments on our proposed benchmark FreeArt-21, Video2Articulation-S dataset, and real-world objects demonstrate that FreeArtGS consistently excels in the free-moving setting while remaining competitive in previous reconstruction settings.
\end{itemize}
\section{Related Works}

\subsection{Articulated Object Reconstruction}

Articulated object reconstruction is a long-standing problem in 3D computer vision and has been widely researched in recent years. 
Feed-forward network-based methods~\cite{ditto,kawana2023detection,qian2022understanding,sun2024opdmulti,geng2023gapartnet,heppert2023carto,yu2024gamma,wang2024rpmart} are trained on an annotated dataset, but fail to generalize to unseen objects. 
To improve the generalization ability, a series of methods leverage the foundation models~\cite{articulateanything,realcode,chen2024urdformer,qiu2025articulate} to generate articulated object models from single-view images. While these approaches benefit from extensive pre-training on diverse datasets, they typically lack geometric consistency constraints and iterative refinement mechanisms. 
Another series regards articulated object reconstruction as calibrated multi-view camera capturing under two distinct articulation configurations~\cite{splart,reartgs,artgs,guo2025articulatedgs,liu2023paris,wang2024sm,weng2024neural}. Despite their geometric rigor, it remains difficult to align the axes of different states, leading to the limited practicality. 
To address these limitations and improve generalization, recently, a new line of methods~\cite{rsrd,video2articulation,wu2025predict,monomobility} reconstruct articulated objects from monocular RGB or RGB-D video sequences under the assumption that one part remains stationary relative to the background. 
These methods suffer from two fundamental limitations. First, the static base part assumption is violated in practical scenarios where users naturally manipulate objects like scissors or pliers. Second, the inability to freely repose the object during capture results in incomplete coverage. In contrast, our method operates in a free-moving setting, where the object pose and joint state can vary concurrently, eliminating these issues.
\vspace{-2pt}
\subsection{Dynamic Reconstruction}
Articulated object reconstruction can be regarded as another type of dynamic reconstruction. Feed-forward methods~\cite{zhang2024monst3r,lu2025align3r,wang2025continuous} directly learn to reconstruct dynamic point clouds from large-scale datasets. However, they fail to recover precise motions under the free-moving setting (Fig.~\ref{fig:vis}). 
Optimization-based dynamic reconstruction methods~\cite{wu20244d,cao2023hexplane,huang2024sc,lin2024gaussian} reconstruct temporal deformations of radiance fields, but often lack generalization ability. Recently, point tracking methods~\cite{alltracker,tapip3d,xiao2025spatialtrackerv2,rajivc2025multi,ngodelta,yao2025uni4d} provide generalized priors at pixel-level resolution, enabling their application to free-moving articulated object reconstruction. However, due to the data-driven nature, the tracks inevitably contain noise and outliers. Our method uses an off-the-shelf point tracking model~\cite{alltracker} to generate pseudo motion labels, and optimizes the articulated object to fit the labels. In this way, we combine the generalization ability of point tracking and the high accuracy of the optimization-based dynamic reconstruction in one framework.
\section{Method}
\begin{figure*}
    \centering
    \includegraphics[width=1.0\linewidth]{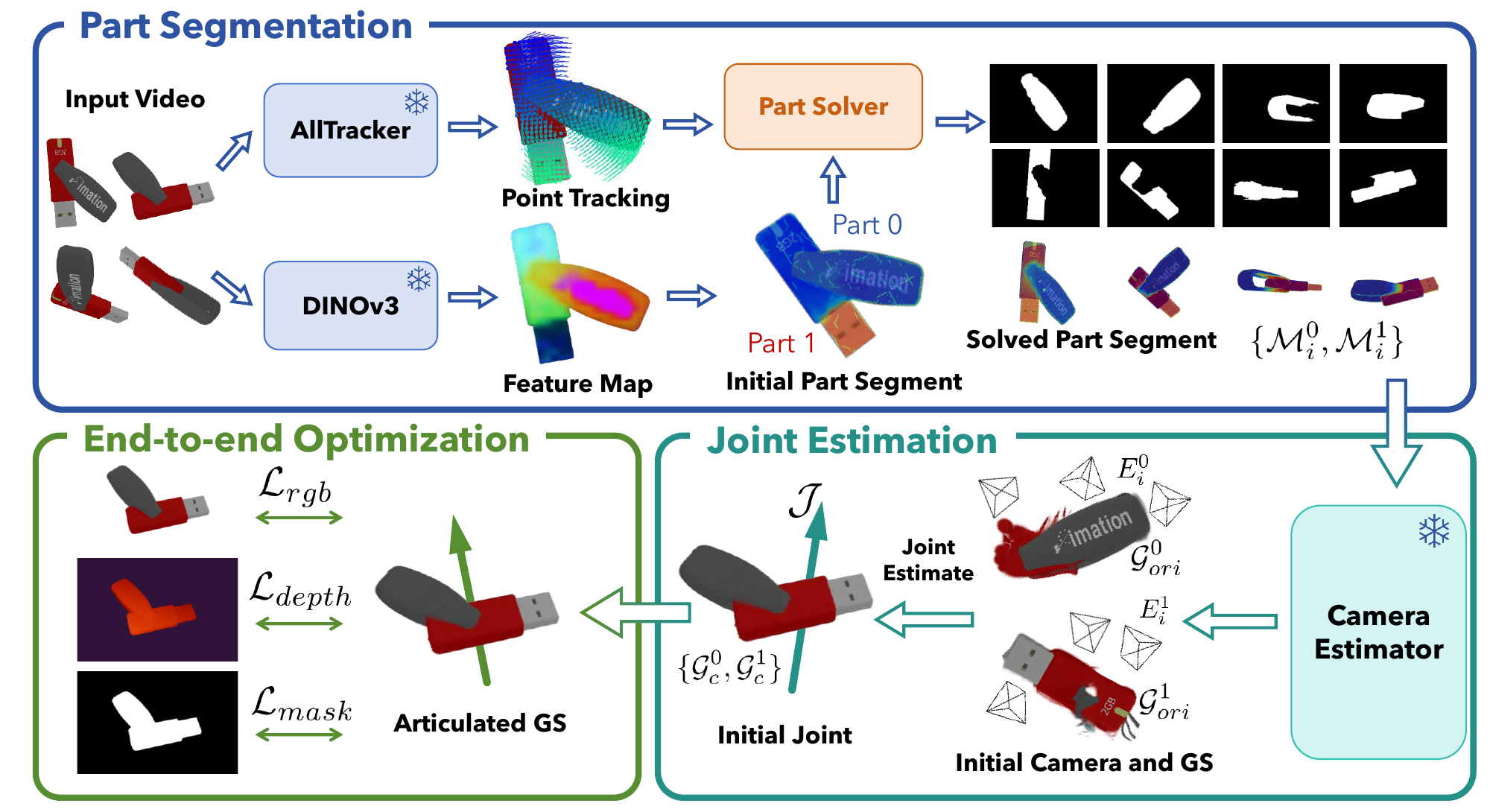}
    \caption{Overview of FreeArtGS. It consists of three modules: (1) Part Segmentation from free-moving video; (2) Joint Estimation with estimated part transforms; (3) End-to-end Optimization for articulated Gaussian Splatting.}
    \label{fig:pipeline}
\end{figure*}
\subsection{Overview}
Given a monocular RGB-D video of 
a free-moving articulated object with two rigid parts $\mathcal{V} = \{\mathcal{I}_i, \mathcal{D}_i\}_{i=1}^N$ and foreground masks $\{\mathcal{M}_i^{fg}\}_{i=1}^N$, obtained using Segment Anything Model~\cite{ravisam}, our goal is to reconstruct its canonical Gaussians $\mathcal{G}_c = \{\mathcal{G}_c^0, \mathcal{G}_c^1\}$ 
of the two parts and joint parameters $\mathcal{J}$. 
Thus, the object can be represented as $\mathcal{G} = \mathcal{G}_c \circ \mathcal{J}$.

As shown in Fig.~\ref{fig:pipeline}, our framework includes three modules: part segmentation, joint estimation, and end-to-end optimization. In Sec.~\ref{sec:seg}, our method leverages the point tracking results of the two parts to obtain their 2D part segmentation $\mathcal{M} = \{\mathcal{M}^0_i, \mathcal{M}^1_i\}_{i=1}^N$ within the foreground masks. In Sec.~\ref{sec:3.3}, we reconstruct the Gaussians of the two parts $\mathcal{G}_{ori}^0$ and $\mathcal{G}_{ori}^1$ respectively, coarsely estimate the joint parameters $\mathcal{J}$, and calibrate $\{\mathcal{G}_{ori}^0, \mathcal{G}_{ori}^1\}$ and $\mathcal{J}$ to the canonical Gaussians $\{\mathcal{G}_c^0, \mathcal{G}_c^1\}$. In Sec.~\ref{sec:opt}, we perform blended rendering and fine-tune the Gaussians and joint parameters with an end-to-end optimization.



\subsection{Free-moving Part Segmentation}
\label{sec:seg}
\noindent\textbf{Setting.} We first aim to segment the articulated object into two rigid parts purely from motion. Our key assumption is that within a short temporal window, the motion of each part is well-approximated by an independent rigid transform. Specifically, for a frame pair $(t, t')$ in the window and a tracked pixel $p$ with valid depth, 
we obtain the corresponding 3D points $X_{t,p}$ and $X_{t',p}$. 
We seek two rigid transforms $T^0_{t\to t'}$ 
and $T^1_{t\to t'}$ and a soft part weight $w_{t,p}\in[0,1]$ per point, 
where $w_{t,p}\!\approx\!1$ and $w_{t,p}\!\approx\!0$ denote the two different parts. We initialize the part weight by clustering with a feature map from DINOv3~\cite{simeoni2025dinov3}, where the semantically similar points are assigned the same part label.

\noindent\textbf{Part Solver.} We process the RGB-D video $\mathcal{V}$ with a sliding window of size $n$. 
For each window, assume $t=0$ for the first frame of the window, AllTracker~\cite{alltracker} provides pixel-level 2D trajectories through $n$ frames
$\{u_{t,p}\}_{t\in[0,n-1]}\in \mathbb{R}^2$, 
where $p\in\mathbb{Z}^2$ is the pixel index of the tracked points in the first frame of the window
and $u_{t,p}$ denotes its 2D position at the frame $t$ in the window.
Then we lift them to 3D trajectories $\{X_{t,p}\}\in \mathbb{R}^3$ with depth and camera intrinsics. 
With these trajectories, we further optimize the rigid transform and part weight $w_{t,p}$ for each part: 
\[
\begin{aligned}
\mathcal{L}_{\mathrm{main}} &= \sum_{p} (1-w_{t,p})\rho\left(\frac{\big\|T^0_{0\to t}X_{0,p}-X_{t,p}\big\|}{\big\|X_{0,p}-X_{t,p}\big\|+\epsilon}\right) \\
&\quad+ w_{t,p}\rho\left(\frac{\big\|T^1_{0\to t}X_{0,p}-X_{t,p}\big\|}{\big\|X_{0,p}-X_{t,p}\big\|+\epsilon}\right),
\end{aligned}
\]
where $\rho(\cdot)$ is Huber loss and $\epsilon$ is a small constant to avoid division by zero.

\noindent\textbf{Regularization.} Jointly optimizing the transform and part weight may fall into a sub-optimal solution. To this end, we regularize $w_{t,p}$ to be confident and spatially coherent. First, an entropy penalty encourages near-binary assignments,
\[
\mathcal{L}_{\mathrm{ent}}=-\sum_{p}\Big[w_{t,p}\log w_{t,p}+(1-w_{t,p})\log(1-w_{t,p})\Big].
\]
Second, to prevent the model from fitting the unstable point tracking results, we build a feature-space neighbor graph $\mathcal{N}(p)$ by sampling per-pixel image features at tracked points and connecting radius-based neighbors with weights $\alpha_{pq}$. A smoothness term enforces local consistency,
\[
\mathcal{L}_{\mathrm{smooth}}=\sum_{p}\sum_{q\in\mathcal{N}(p)}\alpha_{pq}\,\big|w_{t,p}-w_{t,q}\big|.
\]
Finally, we discourage part weights from being too different from the initial weights $w_{0,p}$ with BCE loss $\mathcal{L}_{\mathrm{init}}$,
\[
\mathcal{L}_{\mathrm{init}}=\sum_{p}\mathrm{BCE}\!\left(w_{t,p},\,w_{0,p}\right).
\]
Our objective per window is
\[
\mathcal{L}=\lambda_m\mathcal{L}_{\mathrm{main}}
+\lambda_s\mathcal{L}_{\mathrm{smooth}}
+\lambda_e\mathcal{L}_{\mathrm{ent}}
+\lambda_{\mathrm{init}}\mathcal{L}_{\mathrm{init}}.
\]

\noindent\textbf{Part Segmentation.} We propagate the optimized part weights $\{w_{i,p}\}$ across windows, fill the unobserved pixels via feature-space~\cite{simeoni2025dinov3} neighbors, and obtain the binary part masks $\{\mathcal{M}^0_i,\mathcal{M}^1_i\}_{i=1}^N$ by thresholding the part weights at $0.5$. Refer to the supplementary materials for the details.

\subsection{Joint Estimation}
\label{sec:3.3}
 
\noindent\textbf{Part-level Reconstruction.} With $\mathcal{I}_i$, $\mathcal{D}_i$ and $\{\mathcal{M}_i^k\}_{k\in\{0,1\}}$, where $i$ and $k$ are the frame and part index,
we leverage off-the-shelf pose estimators~\cite{bundlesdf,arun1987least} 
to calibrate each frame's part-to-camera transformations $\{{E}_i^k\}\in \mathrm{SE}(3)$ for each part. We note that, though we have obtained $T^k_{t\to t'}$ in part segmentation, solving the part-to-camera transformations from all the pairs is not trivial~\cite{wang2024dust3r}, since the motion tracking labels $\{u_{t,p}\}$ contain noise and outliers. In contrast, off-the-shelf pose estimators are robust to sudden visual changes while preserving multi-view consistency. With part masks $\{\mathcal{M}_i^k\}$ and transformations $\{{E}_i^k\}$, we reconstruct each part $\mathcal{G}_{ori}^k$ and optimize their poses with 3DGS~\cite{kerbl20233d}.

\noindent\textbf{Pose Calibration.} To unify the object-to-camera poses, we calibrate the poses of two parts to a canonical coordinate system by a rigid transform $A^k \in \mathrm{SE}(3)$, which is given as the inverse of ${E}_0^k$ in the first frame. We choose the part with the least average moving as the reference part (denoted as part 0) and the other part as the relative moving part (denoted as part 1). After calibration, we obtain the canonical Gaussians $\mathcal{G}_c^k = \mathcal{G}_{ori}^k \circ A^k$, transformation $E^{ref}_i = {E}_i^0 \circ {A}^0$ of the reference part and ${E}^{rel}_i = {E}_i^1 \circ {A}^1$ of the relative part. By aligning both parts, their trajectories share the same axes in the first frame, and the relative transformation of ${E}^{ref}_i$ and ${E}^{rel}_i$ represents the combination of joint state and axes.

\noindent\textbf{Joint Type Estimation.} We estimate the kinematic joint that explains the relative transformation between the two reconstructed parts. From $\{{E}^{ref}_i\}$ and $\{{E}^{rel}_i\}$, we obtain a sequence of relative part poses $\{{T}_i\}_{i=1}^N\in\mathrm{SE}(3)$ of one part w.r.t. the other, and implement a light-weight solver that determines the joint type and estimates joint parameters. We decide the joint type by two cues: the overall rotation span across frames and whether translations lie nearly on a single line. A small span with strong linearity indicates a prismatic joint; otherwise, we regard the joint as revolute.

\noindent \textbf{Joint Axis Estimation.} For a \textbf{revolute} joint, we solve the \textit{closed-form rotation axis} from pairwise relative rotations, recover the per-frame angle by fitting pairwise angle differences with the first frame, and solve the pivot only on the plane orthogonal to the axis to avoid degeneracy. For a \textbf{prismatic} joint, we recover the translation axis with \textit{PCA} from $\{{T}_i\}_{i=1}^N$, project each translation onto the axis to obtain the displacement sequence, and keep a constant rotation. 

\noindent \textbf{Noise Resistance.} The part poses estimated inevitably contain noise from the off-the-shelf methods. To ensure robustness, there are two key designs in joint estimation. First, instead of directly estimating the joint from the absolute ${T}_i$, we solve the pairwise relative transform ${T}_{i \to (i+1)}$ between neighboring frames. Second, we filter the outlier transforms with a threshold of $2\sigma$, where $\sigma$ is the standard deviation of the translations of poses in 3D space. These choices make the solver stable under small tracking errors and occasional outlier frames, while remaining fast and light-weight.


\begin{table*}[t]
    \centering
    \caption{Results and Ablation Studies on FreeArt-21. Metrics are reported as mean $\pm$ std over all test videos. Lower ($\downarrow$) is better for the joint and geometry metrics, higher ($\uparrow$) is better for the rendering metric. The best results are highlighted in \textbf{bold}.}
    \label{tab:results_free}
    \renewcommand\tabcolsep{7pt}
    \resizebox{\linewidth}{!}{%
    \begin{tabular}{llcccccccc}
    \toprule
    \textbf{Joint Type} & \textbf{Method} & \textbf{Axis (deg)$\downarrow$} & \textbf{Position (cm)$\downarrow$} & \textbf{State (deg/cm)$\downarrow$} & \textbf{CD-w (cm)$\downarrow$} & \textbf{CD-m (cm)$\downarrow$} & \textbf{CD-s (cm)$\downarrow$} & \textbf{PSNR (dB)$\uparrow$} \\
    \midrule
    \multirow{6}{*}{\centering Revolute} 
    & {ArticulateAnything~\cite{articulateanything}} & 42.00$\pm$46.48 & 59.38$\pm$62.19 & - & - & - & - & - \\
    & {Video2Articulation~\cite{video2articulation}} & 20.00$\pm$28.81 & 16.31$\pm$28.78 & 27.37$\pm$19.06 & 2.29$\pm$3.40 & 10.74$\pm$16.35 & 1.87$\pm$1.59 & - \\
    & Ours w/o Smooth Loss & 28.01$\pm$29.23 & 17.73$\pm$30.79 & 18.74$\pm$20.96 & 5.72$\pm$7.39 & 14.37$\pm$15.89 & 5.64$\pm$9.17 & 10.60$\pm$4.61 \\
    & Ours w/o Init Loss & 9.35$\pm$13.50 & 19.58$\pm$39.17 & 14.64$\pm$19.90 & 0.75$\pm$0.90 & 1.90$\pm$3.11 & 1.14$\pm$2.60 & 13.07$\pm$7.26 \\
    & Ours w/o Noise Resistance & 4.75$\pm$7.83 & 2.22$\pm$6.89 & \textbf{1.30$\pm$1.09} & 0.17$\pm$0.15 & 0.48$\pm$0.79 & 1.10$\pm$2.72 & 22.65$\pm$3.13 \\
    & Ours w/o Blended Rendering & 1.72$\pm$2.38 & 1.88$\pm$6.23 & 1.88$\pm$2.54 & \textbf{0.12$\pm$0.07} & 0.34$\pm$0.52 & 1.05$\pm$2.61 & 22.23$\pm$2.64 \\
    & \textbf{Ours} & \textbf{1.04$\pm$1.03} & 
    \textbf{0.29$\pm$0.36} & 1.43$\pm$1.20 & 0.14$\pm$0.13 & \textbf{0.28$\pm$0.36} & \textbf{0.97$\pm$2.69} & \textbf{24.02$\pm$3.15} \\
    \midrule
    \multirow{6}{*}{\centering Prismatic} 
    & {ArticulateAnything~\cite{articulateanything}} & 45.00$\pm$49.30 & - & - & - & - & - & - \\
    & {Video2Articulation~\cite{video2articulation}} & 18.47$\pm$22.83 & - & 13.98$\pm$21.47 & 1.51$\pm$2.23 & 8.41$\pm$14.57 & 5.31$\pm$15.37 & - \\
    & Ours w/o Smooth Loss & 26.97$\pm$23.75 & - & 46.91$\pm$58.06 & 8.21$\pm$11.16 & 11.82$\pm$11.37 & 10.16$\pm$13.62 & 12.63$\pm$6.46 \\
    & Ours w/o Init Loss & 28.72$\pm$27.83 & - & 38.31$\pm$38.81 & 22.3$\pm$50.18 & 23.73$\pm$44.37 & 21.62$\pm$45.33 & 13.86$\pm$4.48 \\
    & Ours w/o Noise Resistance & 11.90$\pm$26.93 & - & 16.78$\pm$38.86 & 0.87$\pm$1.02 & 5.85$\pm$13.24 & 0.90$\pm$1.50 & 20.18$\pm$3.36 \\
    & Ours w/o Blended Rendering & 2.04$\pm$3.01 & - & 0.97$\pm$0.39 & 0.45$\pm$0.24 & \textbf{0.50$\pm$0.42} & \textbf{0.28$\pm$0.13} & 20.24$\pm$2.64 \\
    & \textbf{Ours} & \textbf{1.85$\pm$2.75} & \textbf{-} & \textbf{0.90$\pm$0.53} & \textbf{0.41$\pm$0.22} & 0.67$\pm$0.34 & 0.30$\pm$0.13 & \textbf{22.92$\pm$2.65} \\
    \bottomrule
    \end{tabular}%
    }
    \end{table*}

\subsection{End-to-end Optimization}
\label{sec:opt}
\noindent\textbf{Joint Formulation.} Starting from an estimated joint on the canonical poses detailed in \ref{sec:3.3}, we perform an axis-aware end-to-end optimization that jointly refines appearance, geometry, cameras, and articulation. Denoting $I$ as the identity matrix, we parameterize the target part by either a revolute joint with unit axis $u$, pivot $o$, and per-frame angle $\theta_i$, or a prismatic joint with axis $u$ and displacement $d_i$:
\[
T_i=
\begin{cases}
T(\theta_i;u,o)
= \big[R(u,\theta_i)\mid(I-R(u,\theta_i))o\big],\\[6pt]
T(d_i;u)
= \big[I\mid d_i u\big].
\end{cases}
\]


\noindent\textbf{Blended Rendering and Optimization.} We apply the rigid part transformations on the canonical Gaussians $\mathcal{G}_c$ for frame $i$, and then perform alpha-blend on the Gaussians according to part weights $w=\{w_i\}\in [0,1]$ which represent the the probability of each Gaussian belonging to both parts
\[\textstyle \mathcal{G}_i = w(\mathcal{G}_c \circ I)\cup(1-w)(\mathcal{G}_c\circ\mathcal{J}_i).\]
Finally, we render frame~$i$ with a differentiable renderer
\[\textstyle \hat{\mathcal{I}}_i=\mathcal{R}\big(\mathcal{G}_i,{K}_i,{E}_i^{ref}\big)\]
where $K_i$ is the camera intrinsics.\\
Supervised with RGB, depth, and foreground masks, we optimize Gaussian parameters for both parts, camera poses, part weights and articulation variables $\{{u},{o},\{\theta_i\}\}$ or $\{{u},\{d_i\}\}$ together. The full objective is
\[\textstyle \mathcal{L}_{E2E}=\sum_i\big(\mathcal{L}_{\mathrm{rgb}}^i+\lambda_{\mathrm{depth}}\,\mathcal{L}_{\mathrm{depth}}^i+\lambda_{\mathrm{mask}}\,\mathcal{L}_{\mathrm{mask}}^i\big)\]
where
\[
\mathcal{L}_{\mathrm{rgb}}^i
=(1-\lambda_{\mathrm{ssim}})\,\mathcal{L}_{1}(\hat{\mathcal{I}}_i, \mathcal{I}_i)
  + \lambda_{\mathrm{ssim}}\,\mathcal{L}_{ssim}(\hat{\mathcal{I}}_i, \mathcal{I}_i)
\]
\[
\mathcal{L}_{\mathrm{depth}}^i
= \left| \hat{\mathcal{D}}_i - \mathcal{D}_i \right|
\]
\[
\mathcal{L}_{\mathrm{mask}}^i
= \left| A_i - {\mathcal{M}}_i \right|
\]
$\mathcal{L}_{1}$ and $\mathcal{L}_{ssim}$ follow the definitions in~\cite{kerbl20233d}. This stage tightly couples appearance with kinematics and corrects small biases from the coarse joint, producing a high-fidelity articulated 3DGS.
   
\section{Experiments}

\subsection{Experimental Settings}
\noindent\textbf{Datasets.} We evaluate the reconstruction performance of our method on the following datasets:
\label{Datasets}

\noindent\textbf{(1) New Benchmark: FreeArt-21.} As there is no existing benchmark for free-moving articulated object reconstruction, to evaluate our method, we propose FreeArt-21, a new benchmark containing 21 objects of 7 different categories (5 revolute and 2 prismatic) from the PartNet-Mobility dataset~\cite{partnetmobility}. To simulate the free-moving setting, we deploy a VR system that is widely used in augmented reality and robotics, place the object in front of a fixed RGB-D camera, and teleoperate the object. Please refer to the supplementary materials for the details of our benchmark.

\noindent\textbf{(2) Video2Articulation-S.} Video2Articulation-S is a synthetic dataset of two-part articulated objects proposed by Video2Articulation~\cite{video2articulation}.
It consists of 73 test videos across 11 categories of synthetic objects from the PartNet-Mobility
dataset, where each object has a static base part. Since a static-base capture can be regarded as a special case of the free-moving setting, our method is also compatible with it.

\noindent\textbf{(3) Real-world Articulated Objects. } We evaluate our method on six daily articulated objects, including five revolute-joint objects and one prismatic-joint object. The free-moving videos are captured while the objects are held by hand. We fix an Orbbec Femto Bolt RGB-D camera, and the object holder stands at a distance of 30-50cm from the camera. We report both qualitative and quantitative results.

\begin{figure*}[t]
    \centering
    \includegraphics[width=1.0\linewidth]{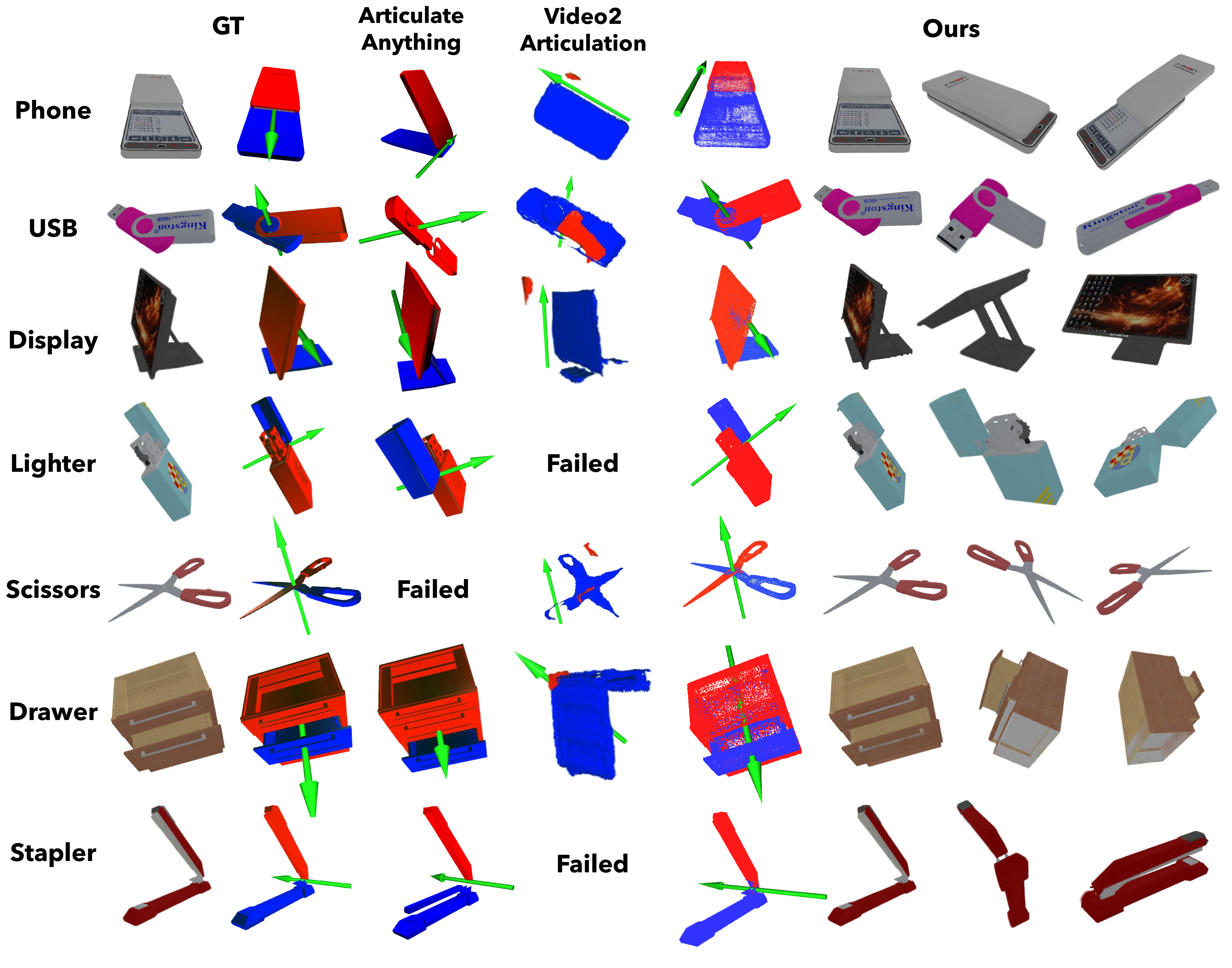}
    \caption{Qualitative Results on FreeArt-21. We visualize both the articulation and rendering results of our method. The red part and the blue part are the identified parts by our method.}
    \label{fig:vis}
\end{figure*}

\noindent\textbf{Evaluation Metrics.}
We report the following core metrics:\\
(1) Joint axis error (degree): both for revolute and prismatic joints, 
the angle between the predicted unit axis ${u}$ and ground truth ${u}_{gt}$,
(2) Joint position error (cm): only for revolute joints, the Euclidean distance between predicted pivot ${o}$ and ground-truth pivot ${o}_{gt}$,
(3) State (degree/cm): both for revolute and prismatic joints, the absolute difference between the predicted joint state and the ground truth joint state.
(4) Chamfer Distance (cm): symmetric $\ell_2$ Chamfer distance between reconstructed and ground-truth surfaces. 
We report CD on the whole object (CD-w) 
and separately on the \emph{moving} part (CD-m) and the \emph{reference} part (CD-s). 
All distances are computed in the canonical state and reported in centimeters.
(5) PSNR: On our FreeArt-21 dataset, 
we additionally report PSNR of novel views and joint states measured inside the foreground mask. 

\noindent\textbf{Baselines.} We choose Video2Articulation~\cite{video2articulation}, Robot-See-Robot-Do (RSRD)~\cite{kerrrobot} and Articulate-Anything~\cite{articulateanything} as our baseline methods. Articulate-Anything is a foundation model-based method that predicts the whole URDF from only a single image input. Since it also uses PartNet-Mobility as the URDF template for retrieval, the domain gap is reduced. For Articulate-Anything, we use the GPT-4o~\cite{hurst2024gpt} model as the vision-language model. Additionally, we provide the ID of each case to the model, which means the model only needs to predict the correct joint, rather than jointly inferring both the object mesh and the joint.
RSRD first reconstructs the object with a smartphone scan and recovers the part motion from a monocular video. Since FreeArt-21 only contains free-moving object frames, we only evaluate its performance on Video2Articulation-S. Video2Articulation leverages the feed-forward point map models to predict the dynamics. Although not the same as the free-moving scenario, it is the latest open-source state-of-the-art method in the setting closest to ours.

\begin{table*}[t]
    \centering
    \caption{Results on Video2Articulation-S Dataset. We report joint estimation results (top) and geometry reconstruction results (bottom). The best results are in \textbf{bold}.}
    \label{tab:combined_results}
    \renewcommand\tabcolsep{11pt}
    \begin{tabular}{ll ccc}
    \toprule
    \textbf{Joint Type} & \textbf{Method} & \textbf{Axis (deg)$\downarrow$} & \textbf{Position (cm)$\downarrow$} & \textbf{State (deg/cm)$\downarrow$} \\
    \midrule
    \multirow{4}{*}{Revolute} 
    & Articulate-Anything~\cite{articulateanything} & 46.98$\pm$45.27 & 81.00$\pm$40.00 & N/A \\
    & RSRD~\cite{rsrd} & 67.06$\pm$29.22 & 203.00$\pm$748.00 & 59.02$\pm$34.38 \\
    & Video2Articulation~\cite{video2articulation} & 18.34$\pm$32.09 & 13.00$\pm$25.00 & 14.32$\pm$26.35 \\
    & \textbf{Ours} & \textbf{1.77$\pm$2.87} & \textbf{1.31$\pm$1.81} & \textbf{3.69$\pm$6.60} \\
    \midrule
    \multirow{4}{*}{Prismatic} 
    & Articulate-Anything~\cite{articulateanything} & 52.71$\pm$44.69 & - & N/A \\
    & RSRD~\cite{rsrd} & 69.91$\pm$24.07 & - & 70.00$\pm$48.00 \\
    & Video2Articulation~\cite{video2articulation} & 13.75$\pm$18.91 & - & 8.00$\pm$22.00 \\
    & \textbf{Ours} & \textbf{0.77$\pm$2.30} & - & \textbf{1.00$\pm$2.19} \\
    
    \midrule \midrule
    \textbf{Task} & \textbf{Method} & \textbf{CD-w (cm)$\downarrow$} & \textbf{CD-m (cm)$\downarrow$} & \textbf{CD-s (cm)$\downarrow$} \\
    \midrule
    \multirow{4}{*}{\begin{tabular}[c]{@{}l@{}}Geometry \\ Reconstruction\end{tabular}}
    & Articulate-Anything~\cite{articulateanything} & 11.00$\pm$22.00 & 59.00$\pm$73.00 & 7.00$\pm$18.00 \\
    & RSRD~\cite{rsrd} & 339.00$\pm$2147.00 & 82.00$\pm$117.00 & 14.00$\pm$41.00 \\
    & Video2Articulation~\cite{video2articulation} & \textbf{1.00$\pm$1.00} & 13.00$\pm$26.00 & 6.00$\pm$19.00 \\
    & \textbf{Ours} & 1.87$\pm$2.19 & \textbf{1.00$\pm$2.22} & \textbf{2.39$\pm$2.50} \\
    \bottomrule
    \end{tabular}
\end{table*}

\subsection{Implementation Details}

For all modules, we maintain a unified set of hyperparameters across all datasets, avoiding per-case tuning.

\noindent\textbf{Part segmentation.} For the optimization process, we employ the Adam optimizer with a learning rate of 1e-4 for the rigid transforms $T^0_{t\to t'}$ and $T^1_{t\to t'}$, and 1e-2 for the part weight $w_{t,p}$. 
A sliding window of 8 frames is defined, within which we optimize frame pairs anchored by the first frame, 
specifically $(0, i)$ for $i \in \{1, 2, \dots, 7\}$. Each pair undergoes 100 iterations of optimization. 
The loss weights are: $\lambda_m=200$, $\lambda_s=10$, $\lambda_{init}=5$, and $\lambda_{e}=0.01$.

\noindent\textbf{Joint estimation and end-to-end optimization.} Our implementation is based on NeRFStudio~\cite{nerfstudio} and its default parameters. During reconstruction, we optimize for 30000 iterations in both part-level reconstruction and end-to-end optimization. In the stage of part-level reconstruction, we choose $\lambda_{depth}=1.0, \lambda_{mask}=0.01$ while in the end-to-end optimization, $\lambda_{depth}=1.0, \lambda_{mask}=1.0$.

\noindent\textbf{Hardware and time cost.} We evaluate the running times of our method and the two baselines on a workstation equipped with an Intel i9-14900K CPU and an NVIDIA RTX 4090 GPU. Given an input RGB-D video with 100 frames and a resolution of 640$\times$360, our method takes $\sim$25 minutes, including 6 minutes for part segmentation, 1 minute for joint estimation, and 18 minutes for end-to-end optimization. 
\subsection{Results on FreeArt-21}

As shown in Table \ref{tab:results_free}, across all the 21 objects in the dataset, our method achieves an average error of around 1 degree in axis angle and less than 1 cm in geometry, surpassing all the baselines. FreeArtGS also achieves a plausible PSNR result, while the two baselines fail to recover the precise visual textures. Since the rendering quality is a synergy of pose estimation and visual reconstruction, incorporating a better pose estimation method may result in a higher PSNR. Figure~\ref{fig:vis} presents qualitative results across all categories. The visualization results demonstrate that our method jointly achieves high fidelity in articulation, geometry, and rendering. On challenging thin objects such as scissors, staplers, phones, and USBs, our method precisely reconstructs the geometry, consistent with Table \ref{tab:results_free}. This indicates the robustness of our part segmentation module. Although using PartNet-Mobility as the asset library, Articulate-Anything~\cite{articulateanything} often fails to predict the correct part and joint axis, likely due to error accumulation in the vision-language reasoning process. Video2Articulation also performs poorly on our dataset, since Monst3R~\cite{zhang2024monst3r} fails to predict the moving part in the free-moving scenario.



\begin{figure*}[t]
    \centering
    \includegraphics[width=1.0\linewidth]{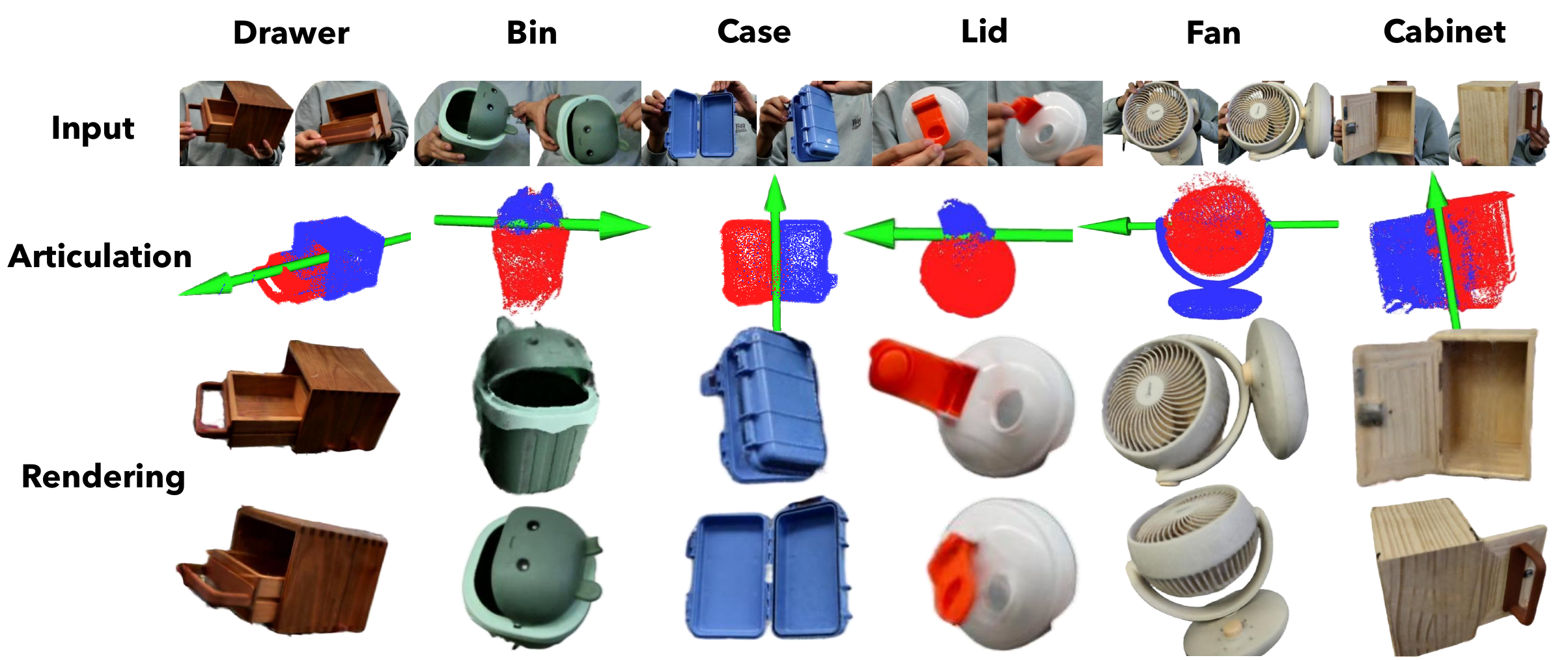}
    \caption{Qualitative Results on Real-world Objects. Our method successfully reconstructs all the objects with correct joints, geometries, and textures.}
    \label{fig:real}
\end{figure*}

\subsection{Results on Video2Articulation-S}

As shown in Table \ref{tab:combined_results}, although under a similar yet different setting, our method also surpasses all baselines on most metrics, consistent with the results on FreeArt-21. RSRD performs worst on all metrics, due to its assumption that the moving patterns of each part are unique, while for articulated objects, their motions are related by the joint transformation. Articulate-Anything also predicts incorrect assets in most cases, likely due to hallucination in the vision-language model. Regarding Video2Articulation, it should be noted that, even under its own setting, the performance of Video2Articulation is still worse than our method. The main reason is that Video2Articulation depends on the predictions from a pretrained feed-forward reconstruction model, which is not robust due to the confidence threshold. Instead, our method only uses the off-the-shelf models as initialization and partial supervision. Combining the priors with optimization is key to the performance gain.

\subsection{Results on Real-world Articulated Objects}

\begin{table}[t]
    \centering
    \caption{Quantitative Results on Real-world Objects. The rotation and translation errors are clipped to $0.1^\circ$ and $0.1$ cm, respectively, which correspond to the smallest annotation units.}
   
    \label{tab:quantitative results of real}
    \renewcommand\tabcolsep{7pt}
    \resizebox{0.48\textwidth}{!}{
    \begin{tabular}{lcccc}
    \toprule
    \textbf{Objects} & \boldmath\textbf{Axis (deg)$\downarrow$} & \boldmath\textbf{Position (cm)$\downarrow$} & \boldmath\textbf{CD (cm)$\downarrow$} & \boldmath\textbf{PSNR (dB)$\uparrow$} \\
    \midrule
    Drawer & 1.06 & - & 1.91 &  19.82  \\
    Bin & 4.68 & 0.20 & 3.46 & 23.75   \\
    Case & 0.10 & 0.10 & 1.89 &  22.77  \\
    Lid & 8.92 & 0.40 & 3.36 & 22.39   \\
    Fan & 1.53 & 1.30 & 1.78 &  21.95  \\
    Cabinet & 0.10 & 3.30 & 2.53 &  23.71  \\
    \textbf{Average} & 2.73 & 1.06 & 2.48 & 22.40\\
    \bottomrule
    \end{tabular}%
    }
\end{table}

We further evaluate FreeArtGS on six real-world objects, including a drawer, a trash bin, a case, a bottle lid, an electric fan, and a cabinet. As shown in Figure~\ref{fig:real} and Table~\ref{tab:quantitative results of real}, our method can not only predict the correct joint type and axis, but also reconstruct precise geometry and textures across all six objects. During the data collection, some areas of the objects are inevitably occluded by human hands. However, as can be seen in the figure, our method is robust to this occlusion. There are two reasons. First, the regularization term of the part segmentation module can resist implausible part weights. Second, the end-to-end optimization from RGB-D images corrects the outlier points. These results highlight the potential of FreeArtGS as a scalable digital twin reconstruction for real-world articulated objects.

\subsection{Ablation Study}

\noindent\textbf{Settings.} To verify the effectiveness of each component in our method, we conduct four ablation studies on the FreeArt-21 dataset. For part segmentation, we ablate both the smoothness loss and the initialization regularization term, denoted as \textbf{w/o Smooth Loss} and \textbf{w/o Init Loss}.
To validate the effectiveness of noise resistance in joint estimation, we remove the outlier filtering and use absolute transforms for initialization, denoted as \textbf{w/o Noise Resistance}.
For the blended rendering in end-to-end optimization, we replace the blending with hard assignment, meaning that the part weights remain fixed at 0 or 1 and will not be refined during optimization, denoted as \textbf{w/o Blended Rendering}.

\noindent\textbf{Results.} The results of the ablation study are shown in Table \ref{tab:results_free}, from which we make the following observations: (1) Smoothness over the neighbor graph and consistency with the initialization are important for both joint and geometry in the ablation. This indicates that although the point tracking model can find the correspondence between neighboring frames, its instability may drive the part solver toward suboptimal solutions. (2) Noise Resistance of joint estimation prevents the joint from overfitting the outlier part transforms, as can be seen from the sudden degradation of the axis angles for both revolute and prismatic joints. (3) Blended Rendering improves the visual rendering quality by around 2 dB. For the few metrics in which removing this module yields slightly better results, the difference in metrics is trivial ($\sim$1mm and $\sim$0.1deg/cm). We include this module since it improves rendering quality while maintaining joint accuracy. This is consistent with its role in refining part weights during end-to-end optimization. 


\section{Conclusion}
In this paper, we propose FreeArtGS, a novel method for reconstructing free-moving articulated objects from monocular RGB-D videos. Our method first segments the free-moving parts by combining an optimization-based method with point-tracking priors. Based on the estimated part segments and transformations, it then infers the joint type and axis by fitting the relative motion between parts. Finally, a 3DGS-based end-to-end optimization jointly refines the joint parameters, geometry, and appearance. Experiments demonstrate the robustness and effectiveness of our method in both simulated and real-world settings. With the growing need to rapidly expand articulated digital twins for augmented reality and robotics, our method provides a promising solution with fewer constraints and higher scalability.

FreeArtGS still has several limitations. First, our method currently assumes a two-part articulated object; extending it to multi-part structures by sequentially capturing each moving part remains an important direction. Second, relying on multiple off-the-shelf priors can lead to cascading error accumulation. A potential solution lies in developing a unified feed-forward model to simultaneously predict joints, poses, geometry, and textures. Third, our framework requires monocular RGB-D input. While extending it to RGB-only sequences by predicting continuous video depth is a natural progression, it currently faces challenges regarding depth accuracy. We leave these as future work.


\section*{Acknowledgements}
\label{sec:ack}
\begin{sloppypar}
This work was supported by the National Natural Science Foundation of China (62136001). We would like to thank Jinghang Wu from Peking University for technical support.
\end{sloppypar}
{
    \small
    \bibliographystyle{ieeenat_fullname}
    \bibliography{main}
}

\clearpage
\maketitlesupplementary
\begin{figure*}[htbp]
    \centering
    \includegraphics[width=1.0\linewidth]{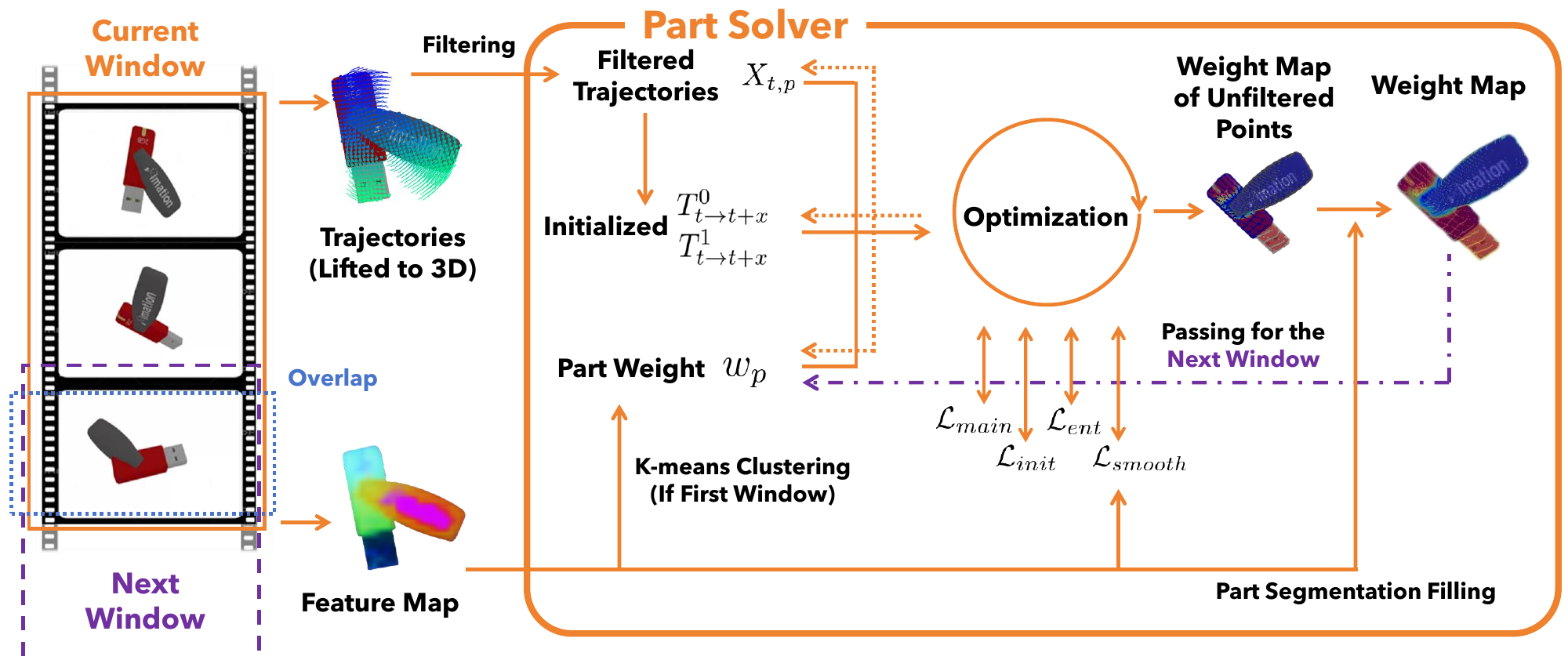}
    \caption{Illustration of Part Solver Module.}
    \label{fig:solver}
\end{figure*}
\section{Method Details}
This section provides additional implementation and methodological details of FreeArtGS:
\begin{itemize}
    \item Sec.~\ref{sec:init}: the initialization procedure of the rigid transformations in part segmentation.
    \item Sec.~\ref{sec:filter}: the trajectory filtering strategy to obtain a reliable subset of trajectories given the point-tracking priors.
    \item Sec.~\ref{sec:partseg}: the propagation algorithm from optimized part weights to segmentations.
    \item Sec.~\ref{sec:camera}: additional description of part pose estimation.
    \item Sec.~\ref{sec:jointtype}: the criterion of deciding the joint type.
    \item Sec.~\ref{sec:jointaxis}: the strategy of estimating the joint axis.
\end{itemize}
Figure~\ref{fig:solver} further illustrates the detailed data flow of the Part Solver module introduced in Sec.~3.2 of the main paper.


\subsection{Transformation Initialization}
\label{sec:init}
As described in Sec. 3.2 of the manuscript, our method processes the input video in a window-wise manner. 
Before each optimization procedure, the rigid transformations must be initialized for the corresponding frame pairs within the window.

For each frame pair $(t, t{+}x)$ with $x \in [1, K-1]$, where $K$ is the window size, let $X_{t,p}$ and $X_{t+x,p}$, with $p\in\mathcal{P}$, denote the 3D positions of trajectory point $p$ at times $t$ and $t{+}x$, respectively. 
Given an initial partition of the trajectories into two parts, $\mathcal{P}_0$ and $\mathcal{P}_1$, we estimate the rigid transforms $T^0_{t\to t+x}, T^1_{t\to t+x} \in \mathrm{SE}(3)$ by minimizing a robust registration objective:
\begin{equation}
    T^k_{t\to t+x}
    = \arg\min_{T \in \mathrm{SE}(3)} 
    \sum_{p \in \mathcal{P}_k} 
    \rho\big(\,\lVert X_{t+x,p} - T X_{t,p} \rVert_2^2\big),
    \label{eq:init_reg}
\end{equation}
where $\rho(\cdot)$ denotes the Tukey loss. 
In practice, we realize this step with a RANSAC-based estimator followed by refinement under the Tukey weighting scheme, which increases robustness to outliers.

To further improve the initialization, we perform an EM-style iterative refinement over trajectory assignments and transforms. 
Let $z_p \in \{0,1\}$ denote the part label of trajectory $p$. 
At iteration $i$, the E-step assigns each trajectory to the transform with the smaller reconstruction error:
\begin{equation*}
    z_p^{(i+1)} 
    = \arg\min_{k \in \{0,1\}} 
    \lVert X_{t+x,p} - T^{k,(i)}_{t\to t+x} X_{t,p} \rVert_2^2
    \label{eq:em_estep}
\end{equation*}
In the M-step, the transforms are re-estimated given the updated assignments by solving
\begin{equation*}
    T^{k,(i+1)}_{t\to t+x} 
    = \arg\min_{T} 
    \sum_{p : z_p^{(i+1)} = k} 
    \rho\big(\,\lVert X_{t+x,p} - T X_{t,p} \rVert_2^2\big)
    \label{eq:em_mstep}
\end{equation*}
The final estimates are used as the initial transformations $T^0_{t\to t+x}$ and $T^1_{t\to t+x}$ for frame pair $(t,t{+}x)$, providing a robust starting point that reduces the risk of convergence to poor local optima.

\subsection{Trajectory Filtering}
\label{sec:filter}
The off-the-shelf point tracking model~\cite{alltracker} can produce noisy trajectories, and directly using all trajectories $\{X_{t,p}\}$ is suboptimal. We therefore apply a multi-stage filtering scheme to obtain a reliable subset of trajectories.

We first retain trajectories that remain visible and inside the foreground masks across frames. 
Let $c_{t,p}$ denote the visibility confidence of trajectory $p$ at time $t$ predicted by the point tracking model, and let $m_{t,p} \in \{0,1\}$ indicate whether the corresponding pixel lies inside the foreground mask at time $t$. 
We define the visibility- and mask-consistent set as
\begin{equation}
    \mathcal{S}_{\mathrm{vis}}
    = \big\{ (t,p) \,\big|\,
        c_{t,p} > \tau_c,\;
        m_{t,p} = 1,\;
        m_{t+1,p} = 1
      \big\},
\end{equation}
where $\tau_c=0.5$ is a visibility threshold.

To further suppress motion outliers, we filter trajectories by displacement magnitude. 
Let $\Delta X_{t,p} = X_{t+1,p} - X_{t,p}$ and denote the mean and standard deviation of $\lVert \Delta X_{t,p} \rVert_2$ over $(t,p) \in \mathcal{S}_{\mathrm{vis}}$ as $\mu_v$ and $\sigma_v$, respectively. 
We keep
\begin{equation*}
    \mathcal{S}_{\mathrm{final}}
    = \big\{ (t,p) \in \mathcal{S}_{\mathrm{vis}} \,\big|\,
        \lVert \Delta X_{t,p} \rVert_2 
        \le \mu_v + \tau_v \sigma_v
      \big\},
\end{equation*}
where $\tau_v=2$ is a displacement threshold. 

\subsection{Part Segmentation}
\label{sec:partseg}
To propagate the part weights from sparse trajectories to the full-pixel map, we leverage DINO features for semantic-aware interpolation. 
Let $\phi_t(u)$ denote the DINO feature at pixel $u$ in frame $t$, and let $\mathcal{U}_t$ be the set of foreground pixels that are not directly covered by the retained trajectories. 
For each $u \in \mathcal{U}_t$, we first compute the cosine similarity between $u$ and each trajectory point:
\begin{equation*}
    s_t(u,p)
    = \cos\!\big(\phi_t(u), \phi_t(p)\big)
    = \frac{\phi_t(u)^\top \phi_t(p)}
           {\lVert \phi_t(u) \rVert_2 \,\lVert \phi_t(p) \rVert_2},
\end{equation*}
The interpolated part weight at pixel $u$ is then defined as the similarity-weighted average
\begin{equation*}
    \tilde{w}_t(u)
    = \frac{
        \sum_p
        s_t(u,p)\, w_{t,p}
    }{
        \sum_p
        s_t(u,p)
    }.
\end{equation*}
The resulting dense part weight map $\tilde{w}_t$ is passed to the next window.

\subsection{Part Pose Estimation}
\label{sec:camera}
As previously discussed, we adopt an off-the-shelf approach to estimate part-to-camera poses, exemplified by BundleSDF~\cite{bundlesdf}. For improved accuracy on synthetic datasets, we replace the feature-matching stage in the Coarse Pose Initialization module of the BundleSDF pipeline with an ICP-based procedure. This adjustment is motivated by the fact that objects in simulation typically exhibit extremely limited texture—many are nearly monochromatic—rendering feature matching unreliable.
For real-world objects, however, we retain the original BundleSDF pipeline unchanged, as it demonstrates robust performance under such conditions.
\subsection{Joint Type Estimation}
\label{sec:jointtype}
We propose a simple criterion to decide whether a motion sequence
should be modeled as a revolute joint or a prismatic joint, based only on
the observed part poses. The decision is made using two scalar features: \textbf{Rotation Amplitude} and \textbf{Translation Linearity Ratio}. We describe the decision criterion as follows.

\paragraph{Input.}
We are given a sequence of relative part poses
\[
T_i \in SE(3), \quad i = 1,\dots,N,
\]
where
\(
T_i =
\begin{bmatrix}
R_i & t_i \\
0 & 1
\end{bmatrix},
\; R_i \in SO(3),\; t_i \in \mathbb{R}^3.
\)
Each $R_i$ is projected to $SO(3)$ to remove numerical noise.

\paragraph{Rotation Amplitude.}
We first measure how much the object rotates over the whole sequence.

We compute the mean rotation by projecting the sum of rotations to
$SO(3)$:
\[
S = \sum_{i=1}^N R_i, \quad
S = U \Sigma V^\top, \quad
R_{\mathrm{mean}} = U V^\top \in SO(3).
\]
For each frame, we form the relative rotation to the mean,
\[
R_i^{\mathrm{rel}} = \operatorname{Proj}_{SO(3)}\!\bigl(R_i R_{\mathrm{mean}}^\top\bigr),
\]
and convert it to an angle
\[
\theta_i = \arccos\!\Bigl(\tfrac{1}{2}(\operatorname{tr}(R_i^{\mathrm{rel}})-1)\Bigr).
\]

To obtain a robust rotation span, we apply an IQR-based outlier
rejection to the set $\{\theta_i\}$ and then define
\[
\Delta\theta = 
\max_{\theta \,\text{inliers}} \theta \;-\;
\min_{\theta \,\text{inliers}} \theta,
\]
and convert it to degrees
\[
\Delta\theta_{\mathrm{deg}} = \frac{180}{\pi}\,\Delta\theta.
\]
This scalar $\Delta\theta_{\mathrm{deg}}$ measures the overall
\textit{rotation amplitude} of the sequence.

\paragraph{Translation Linearity Ratio.}
Independently, we analyze how linear the translation trajectory is.
We first center the translations,
\[
\bar{t} = \frac{1}{N}\sum_{i=1}^N t_i, \quad
x_i = t_i - \bar{t},
\]
and build the (unnormalized) covariance matrix
\[
C = \sum_{i=1}^N x_i x_i^\top \in \mathbb{R}^{3\times 3}.
\]
Let its eigenvalues in descending order be
\[
\lambda_1 \ge \lambda_2 \ge \lambda_3 \ge 0.
\]
We define a \emph{translation linearity ratio}
\[
\rho = \frac{\lambda_2 + \lambda_3}{\lambda_1 + \varepsilon},
\]
with a small $\varepsilon > 0$ to avoid division by zero.
When the translations lie close to a single straight line, $\lambda_1$
dominates and $\rho$ becomes small.

\paragraph{Decision Criterion.}
Using these two scalar features, we classify the joint type according to
\[
\text{model} =
\begin{cases}
\text{prismatic}, 
& \text{if } \Delta\theta_{\mathrm{deg}} < \theta_{\mathrm{th}}
\;\; \text{or} \;\; \rho < \rho_{\mathrm{th}},\\[4pt]
\text{revolute},
& \text{otherwise}.
\end{cases}
\]
In our implementation, we set
\(
\theta_{\mathrm{th}} = 10^\circ
\)
and
\(
\rho_{\mathrm{th}} = 0.05.
\)

Intuitively, if the object barely rotates (small
$\Delta\theta_{\mathrm{deg}}$), or if its translation is very close
to a straight line in 3D (small $\rho$), the motion is better
explained by a prismatic joint; otherwise we adopt a revolute-joint
model.

\begin{table*}[t]

\centering
\caption{Results of Robustness Analysis.}
\label{tab:more_ablation}
\renewcommand\tabcolsep{7pt}
\resizebox{\linewidth}{!}{%
\begin{tabular}{llcccccccc}
\toprule
\textbf{Joint Type} & \textbf{Method} & \boldmath\textbf{Axis(deg)$\downarrow$} & \boldmath\textbf{Position(cm)$\downarrow$} & \boldmath\textbf{State(deg/cm)$\downarrow$} & \boldmath\textbf{CD-w(cm)$\downarrow$} & \boldmath\textbf{CD-m(cm)$\downarrow$} & \boldmath\textbf{CD-s(cm)$\downarrow$} & \boldmath\textbf{PSNR(dB)$\uparrow$} \\
\midrule
\multirow{4}{*}{\centering Revolute} 
& 2\% Depth Noise & 1.41$\pm$1.49 & 0.51$\pm$0.58 & 1.22$\pm$1.27 & 0.23$\pm$0.30 & 0.30$\pm$0.27 & 1.05$\pm$2.47 & 23.52$\pm$4.65 \\
& AllTracker $\to$ CoTracker3& \textbf{0.88$\pm$0.84} & 0.46$\pm$0.86 & 1.68$\pm$1.30 & 0.79$\pm$2.19 & 0.80$\pm$1.08 & 1.47$\pm$2.43 & 23.87$\pm$4.94 \\
& DINOv3 $\to$ DINOv2 & 1.64$\pm$1.22 & 0.32$\pm$0.47 & \textbf{1.21$\pm$0.77} & \textbf{0.11$\pm$0.11} & \textbf{0.28$\pm$0.27} & 1.04$\pm$2.52 & 23.74$\pm$4.41 \\
& \textbf{Ours} & 1.04$\pm$1.03 & 
\textbf{0.29$\pm$0.36} & 1.43$\pm$1.20 & 0.14$\pm$0.13 & 0.28$\pm$0.36 & \textbf{0.97$\pm$2.69} & \textbf{24.02$\pm$3.15} \\
\midrule
\multirow{4}{*}{\centering Prismatic} 
& 2\% Depth Noise & 1.86$\pm$2.51 & - & 0.92$\pm$0.66 & \textbf{0.34$\pm$0.13} & 1.07$\pm$1.18 & 1.78$\pm$3.03 & 21.75$\pm$2.99 \\
& AllTracker $\to$ CoTracker3 & 1.98$\pm$3.09 & - & 1.12$\pm$0.68 & 0.79$\pm$1.08 & 0.98$\pm$1.22 & 0.94$\pm$1.18 & 21.85$\pm$3.64 \\ 
& DINOv3 $\to$ DINOv2 & \textbf{1.46$\pm$1.63} & - & 0.96$\pm$1.59 & 0.56$\pm$0.40 & 0.99$\pm$0.93 & 0.46$\pm$0.51 & 22.39$\pm$3.92 \\
& \textbf{Ours} & 1.85$\pm$2.75 & \textbf{-} & \textbf{0.90$\pm$0.53} & 0.41$\pm$0.22 & \textbf{0.67$\pm$0.34} & \textbf{0.30$\pm$0.13} & \textbf{22.92$\pm$2.65} \\
\bottomrule
\end{tabular}%
}
\end{table*}

\subsection{Joint Axis Estimation}
\label{sec:jointaxis}
Given a sequence of relative part poses 
\[
    T_i = 
    \begin{bmatrix}
        R_i & t_i\\
        0 & 1
    \end{bmatrix}
    \in SE(3), \quad i=1,\dots,N,
\]
our goal is to estimate the underlying joint axis.
Depending on the joint type inferred in the previous stage, we adopt two different strategies, described below.

\paragraph{Revolute Model.}
For a fixed-axis (revolute) joint, all rotations share a common axis direction.
Let 
\[
    R_{ij} = \operatorname{Proj}_{SO(3)}(R_i R_j^\top)
\]
denote the relative rotation between frames $i$ and $j$, where $\operatorname{Proj}_{SO(3)}$ removes numerical drift.
In an ideal revolute motion, the unknown axis direction $u$ satisfies
\[
    R_{ij} u = u
    \quad\Longleftrightarrow\quad
    (R_{ij}-I)\,u = 0
    \quad\forall\, i<j.
\]
Thus $u$ lies in the approximate null space shared by all matrices $(R_{ij}-I)$.
We therefore minimize the quadratic error
\[
    u^\star = 
    \arg\min_{\|u\|=1}
    \sum_{i<j}
        \|(R_{ij} - I)u\|^2.
\]
Expanding the objective yields the symmetric matrix
\[
    A = 
    \sum_{i<j}
        (R_{ij}-I)^\top (R_{ij}-I),
\]
whose eigenvector associated with the smallest eigenvalue gives the optimal axis direction,
\[
    A u^\star = \lambda_{\min} u^\star,
    \qquad \|u^\star\| = 1.
\]

Once the axis direction is known, the axis point $p$ is estimated from the translational consistency equation for a rigid rotation without screw pitch:
\[
    t_i \approx c + (I - R_i)p.
\]
Subtracting two frames removes the global offset $c$:
\[
    t_i - t_j \approx (R_j - R_i)p.
\]
Since the component of $p$ along $u$ is unobservable, we solve $p$ in the plane orthogonal to $u$.  
Let $P = I - u u^\top$ be the projection onto $u^\perp$, then
\[
    P(t_i - t_j) \approx P(R_j - R_i)p.
\]
Stacking all such constraints yields a least-squares system, whose solution is projected back onto $u^\perp$ to enforce $u^\top p = 0$, giving a unique axis point closest to the origin.

\paragraph{Prismatic Model.}
For a prismatic joint, the rigid body exhibits negligible rotation while its translations $\{t_i\}$ lie approximately on a straight 3D line.
We estimate the sliding direction and axis point from the translational trajectory.

Translations are first centered:
\[
    \bar t = \frac{1}{N}\sum_{i=1}^N t_i,
    \qquad
    x_i = t_i - \bar t.
\]
The covariance matrix
\[
    C = \sum_{i=1}^N x_i x_i^\top
\]
is decomposed as
\[
    C v_k = \lambda_k v_k,
    \qquad
    \lambda_1 \ge \lambda_2 \ge \lambda_3 \ge 0.
\]
If the motion is prismatic, $\lambda_1$ dominates, and the first principal component encodes the translation line direction.  
Thus the prismatic axis is
\[
    w^\star = \frac{v_1}{\|v_1\|}.
\]

The axis point $p_0$ is the closest point on this line to the origin, obtained by projecting the mean translation onto $w^\star$:
\[
    p_0 
    = \bar t - (\bar t^\top w^\star)\, w^\star.
\]
Each frame’s joint displacement is then the signed projection
\[
    d_i = (t_i - p_0)^\top w^\star,
\]
which, together with a constant rotation offset, fully characterizes the prismatic motion.

\section{Benchmark Details of FreeArt-21}

We provide further details about the FreeArt-21 benchmark in this section. All data are collected through tele-operation within the Sapien simulator~\cite{partnetmobility}. A human operator simultaneously controls the 6-DoF poses and joint parameters of articulated objects using a PICO 4 Ultra VR headset and controllers, enabling real-time manipulation and interaction with the objects. The camera is positioned approximately 3 meters away from the object in simulation, providing a realistic perspective for the data collection. Each case in the benchmark contains multi-modal data, including RGB images, depth maps, ground-truth object and part masks, as well as the 6-DoF object poses and joint parameters. These elements provide rich supervision for training and evaluating object pose estimation and articulated manipulation tasks. The data collection involves between 200 and 400 frames per sequence, capturing diverse interaction patterns. 

\begin{figure*}[t]
    \centering
    \includegraphics[width=1\linewidth]{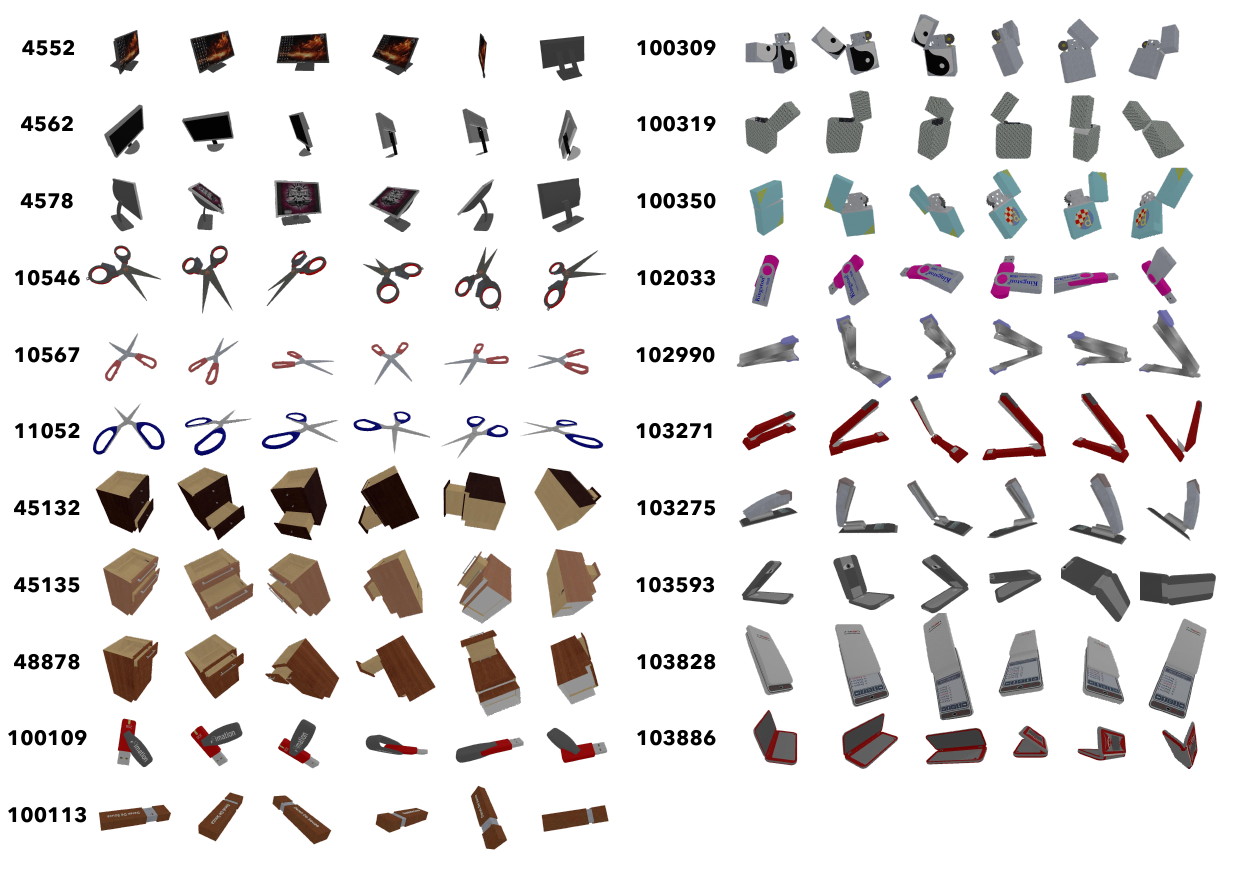}
    \caption{Visualization of all objects in FreeArt-21. }
    \label{fig:benchmark}
\end{figure*}

Compared to other datasets, FreeArt-21 offers more human-like data collection, as the sequences are captured through human teleoperation, rather than automated generation or pre-set trajectories, making it more representative of real-world scenarios where objects are scanned by a person. The objects included in FreeArt-21 are shown in Figure~\ref{fig:benchmark}.

\section{Additional Experiment Results}

\subsection{Robustness Analysis}
We further evaluate the robustness of our framework by replacing AllTracker with CoTracker3 and DINOv3 with DINOv2, while also injecting 2\% Gaussian noise into the depth of the FreeArt-21 dataset. As shown in Table~\ref{tab:more_ablation}, the results indicate that FreeArtGS remains robust to noisy depth inputs and variations in predictions from upstream models.

\subsection{Camera Pose Estimation Results}
\begin{figure}[htbp]
    \centering
    \includegraphics[width=1\linewidth]{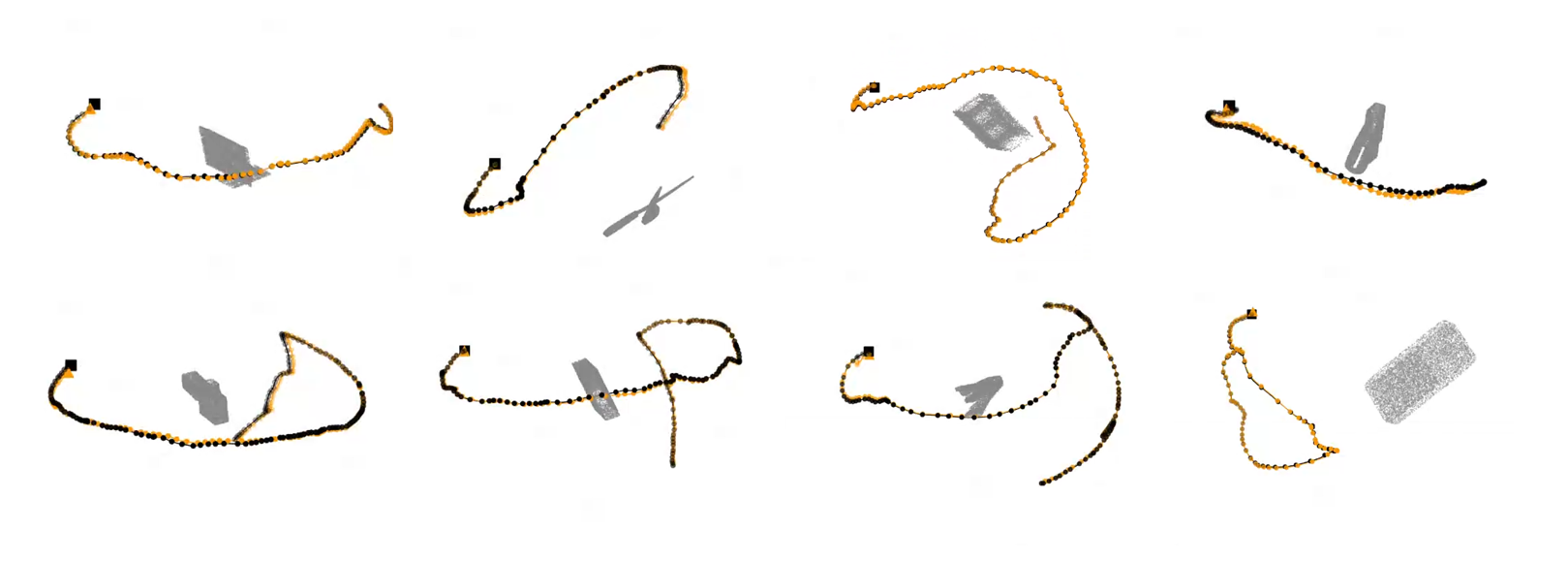}
    \caption{Visualization of camera pose trajectories in FreeArt-21. The black line shows the ground truth camera trajectory, and the yellow line shows our estimated camera trajectory. }
    \label{fig:camera pose vis}
\end{figure}

We evaluate our camera pose estimation results on the Video2Articulation-S dataset and compare them with the baseline method, Video2Articulation~\cite{video2articulation}. Camera pose is a crucial variable that is jointly optimized in our end-to-end pipeline. As shown in Table~\ref{tab:results_video2articulation_cam}, our method achieves significantly lower rotation and translation errors than the baseline, resulting in a more accurate reconstruction of the articulated objects.
We also provide trajectory visualizations in Figure~\ref{fig:camera pose vis} to qualitatively show the alignment between estimated and ground-truth trajectories.

\begin{table}[h]
    \centering
    \caption{Results of Camera Pose Estimation on the Video2Articulation-S Dataset. * indicates that the results are taken from the original paper.}
    \label{tab:results_video2articulation_cam}
    \renewcommand\tabcolsep{7pt}
    \resizebox{\linewidth}{!}{
    \begin{tabular}{lcc}
    \toprule
    \textbf{Method} & \boldmath\textbf{Rot. Error (deg)$\downarrow$} & \boldmath\textbf{Pos. Error (cm)$\downarrow$} \\
    \midrule
    {Video2Articulation*~\cite{video2articulation}} & 4.621 & 9.246 \\
    \textbf{Ours} & \textbf{2.483} & \textbf{1.526} \\
    \bottomrule
    \end{tabular}%
}
\end{table}

\subsection{Part Segmentation Results}
We additionally report the evaluation of our part segmentation. The metrics are calculated in terms of mIoU of the two parts in Table~\ref{tab:iou}. This result confirms that our model produces consistent part segmentations that align closely with the annotated masks.

\begin{table}[h]
    \centering
    \caption{Results of Part Segmentation. * indicates that the results are taken from the original paper.}
    \label{tab:iou}
    \renewcommand\tabcolsep{7pt}
    \resizebox{\linewidth}{!}{
    \begin{tabular}{lcc}
    \toprule
    \textbf{Method / mIoU (\%)$\uparrow$} & \textbf{FreeArt-21} & \textbf{Video2Articulation-S} \\
    \midrule
    {Video2Articulation~\cite{video2articulation}} & 34.61 & 59.65* \\
    \textbf{Ours} & \textbf{90.88} & \textbf{84.92} \\
    \bottomrule
    \end{tabular}%
}
\end{table}

\subsection{More Visualization on FreeArt-21}
\begin{figure*}[b]
    \centering
    \includegraphics[width=1.0\linewidth]{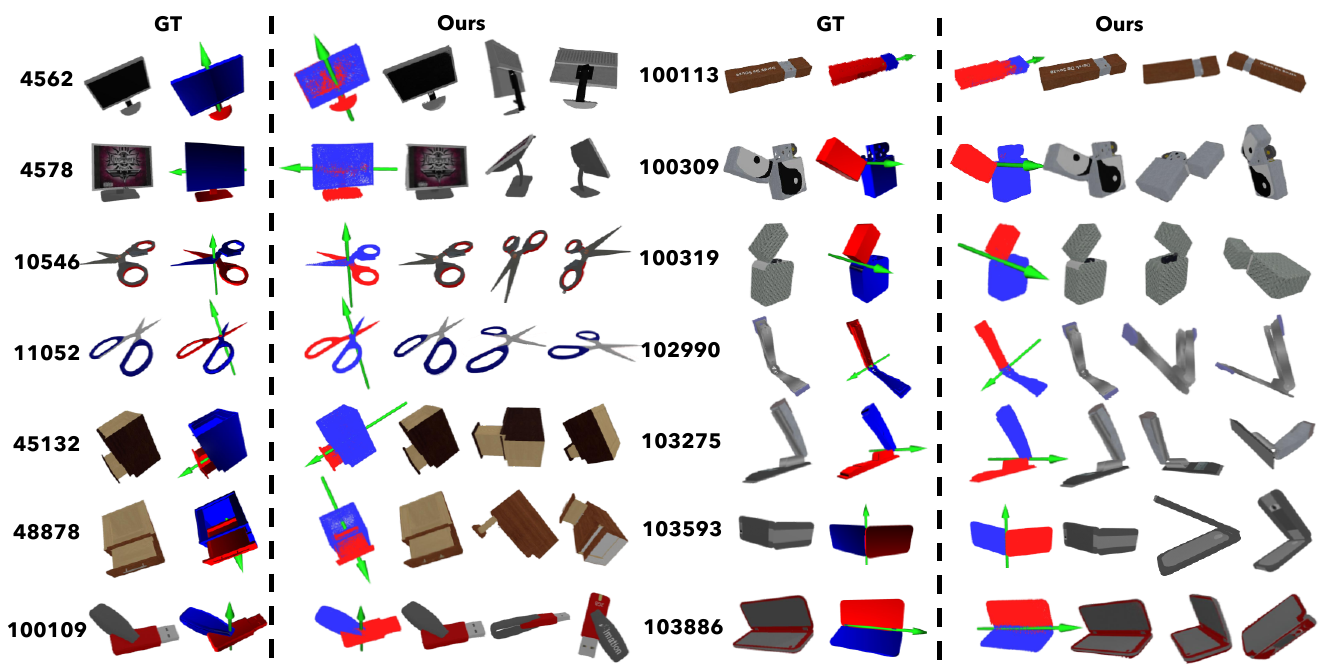}
    \caption{Visualization of our reconstruction results on FreeArt-21.}
    \label{fig:all_vis}
\end{figure*}

We visualize more reconstruction results on FreeArt-21 in Figure~\ref{fig:all_vis}. 

\section{Failure Case Analysis}

\noindent\textbf{Part Segmentation Failures.} Our pipeline's efficacy is closely tied to the accuracy of part segmentation. In cases involving narrow or elongated structures or objects with extremely low texture, the tracked 3D trajectories often exhibit significant drift. These deviations potentially bias the motion-based clustering toward incorrect part assignments. Such mis-segmentation directly undermines the subsequent joint estimation and geometry reconstruction.

\noindent\textbf{Part Pose Ambiguities.} The precision of articulation axis estimation and surface reconstruction is also highly sensitive to part pose accuracy. While we initialize poses using off-the-shelf methods and refine them during optimization, thin or planar objects (e.g., monitors or scissors) present significant challenges. The scarcity of reliable feature correspondences on these geometries often leads to poorly constrained pose optimization. Consequently, the resulting articulation axes may deviate from the physical ground truth, and the reconstructed geometry may suffer from noticeable structural artifacts.



\end{document}